\documentclass[letterpaper, 10 pt, conference]{ieeeconf}
\usepackage{graphicx}
\usepackage{svg}
\usepackage{placeins}
\usepackage{cite}

\IEEEoverridecommandlockouts                              

\title{\LARGE \bf
Comparing Utility of Inertial, Occupancy, Semantic, and Intent Information in Human Motion Prediction During Daily Tasks
}

\author{Max Burns$^{1}$, Maisha Khanum$^{1}$, Monroe Kennedy III$^{1}$ and Steven H. Collins$^{1}$
\thanks{This work has been submitted to the IEEE for possible publication. Copyright may be transferred without notice, after which this version may no longer be accessible.}
\thanks{*This work was supported by the Stanford Graduate Fellowship, the NSF Graduate Research Fellowship, and by Wu Tsai Human Performance Alliance at Stanford University.}
\thanks{$^{1}$Department of Mechanical Engineering, Stanford University, Stanford CA, USA.}%
\thanks{Email: maxburns@stanford.edu}%
}

\begin{document}

\maketitle
\thispagestyle{empty}
\pagestyle{empty}

\begin{abstract}

Accurate human motion prediction is crucial for robotic systems operating around people, particularly in complex indoor spaces. In this study, we assess the relative importance of different sources of information in indoor motion prediction with a human motion diffusion model. We collected a dataset of nine na\"{i}ve human subjects conducting simulated indoor daily activities while wearing a pair of Meta Aria glasses. This dataset includes ten buildings from a university campus, and encompasses 238 minutes of navigation between daily tasks. Overall, we demonstrate a 42\% improvement beyond a constant velocity baseline. Including body motion, scene representation, and eye gaze fixation data significantly reduced prediction error. Semantic information was found to be useful for indoor motion prediction, but to a lesser degree than in outdoor navigation. Providing explicit intent information reduced error beyond any other addition, suggesting that incorporating explicit intent estimation or user input are fundamental for finer prediction of indoor motion. One of the few indicators of intent, eye gaze fixation, was found to be especially useful in predicting deceleration, and provided basic spatial information to the model in the absence of an occupancy map. These results are a first step towards predicting human motion in highly ambiguous indoor scenarios. Code will be made public upon acceptance. Project page: https://human-motion-diffusion.github.io/

\end{abstract}

\section{INTRODUCTION}

Human motion prediction is fundamental to designing effective collaborative robots. For autonomous vehicles, accurately predicting the motion of pedestrians improves safety and can prevent collisions. In-home robots and other systems which work closely in human spaces must avoid both colliding with and obstructing the path of human collaborators. Assistive exoskeletons and prosthetics may also benefit from improved anticipation of how the user will move through a space. Prior work has generally focused on validating different models used to predict future human motion, but there is not yet an established understanding of the relative utility of different information. Particularly, the benefits of increasing system complexity to incorporate different forms of information is poorly understood. Significant prior work focuses on predicting human motion in crowd scenarios or outdoor spaces, due to relevance for autonomous vehicles. Indoor motion is understudied despite occurring commonly in daily life, and is challenging due to frequent speed and direction changes. 

Rather than focusing on comparison of different model structures, we contribute an assessment of the indoor motion prediction problem, in addition to demonstrating a simple but flexible diffusion model structure. These findings are based upon a diverse dataset of nine na\"{i}ve participants completing simulated daily tasks inside spaces on a university campus. 

\begin{figure}[t]
    \centering
    \includegraphics[width=\columnwidth]{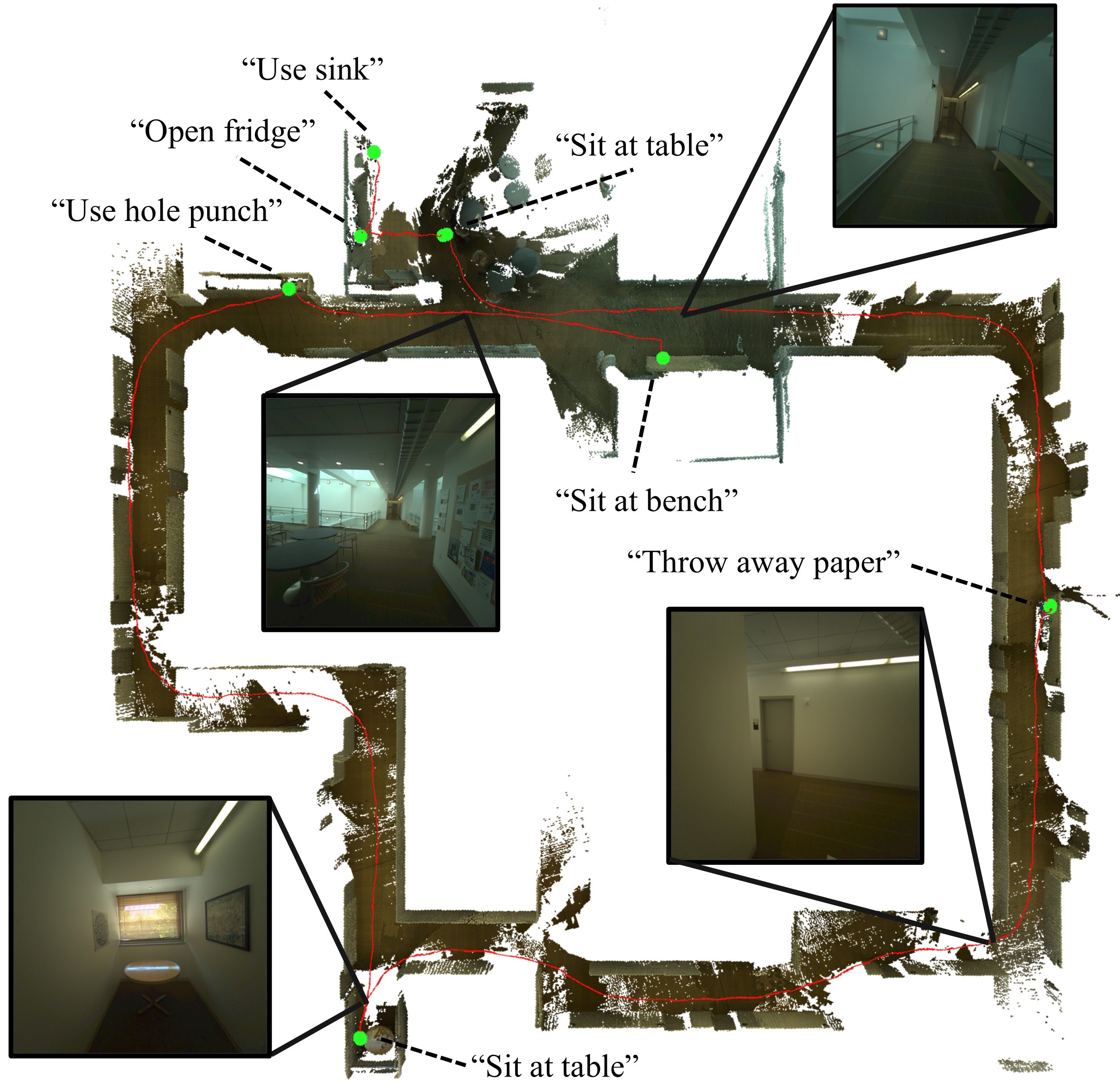}
    \caption{Example task set and reconstruction from protocol. Paths taken by the subject are shown in red, tasks are labeled in green with explanations. Example frames of egocentric data used for the 3D reconstruction are included.}
    \label{fig:top_down_map}
\end{figure}

\section{RELATED WORK}

\subsection{Information in Human Motion Prediction}

Quantifying the benefits of including information in path prediction is challenging due to the lack of established metrics, datasets, or ablation approaches. Thus, we summarize percentage improvements from the inclusion of further information to approximate prior observed benefits.

Prior work in human motion prediction has focused on improving model structures to extract relevant information in making predictions. Early efforts leveraged sequences of trajectory points to forecast future motion, using the evolution of an individual's movement to predict the next few steps \cite{Salzmann2020}. The path a human takes and the speed at which they walk also tend to minimize the cumulative metabolic cost to reach the destination \cite{Brown2021}. Whether this holds in typical real-world walking has yet to be examined explicitly, particularly in cases with other competing objectives.

The incorporation of the scene around an individual has seen significant investigation in recent years due to greater maturity and availability of tools to extract high density 3D scene structure and semantic encoding. LookOut observed a 34\% error reduction from including a 3D version of a scene (compared to 2D) \cite{Pan2025}. TrajForecast also observed a 34\% decrease in error when incorporating depth data, compared to just past motion \cite{Qiu2022}. Including a 3D representation reduced error by a more modest 2\% in EgoNav, when compared to a model based upon only past motion \cite{Wang2026}.

Pretrained vision models have been critical to perform detailed feature extraction from images, and allow lighter predictive models with less training data to incorporate greater environmental context than would be possible from an end-to-end approach \cite{Zheng2022, Escobar2024, Qiu2025}. DINO (self-Distillation with NO labels) is a frequently used backbone to process an image into a usable latent space, which provides a feature space which is easier to reason over than color pixels \cite{Pan2025, Wang2026, Qiu2025}. Some work has also incorporated explicit task-relevant semantic labels through text-based segmentation models, like SAM3 (Segment Anything Model) \cite{Carion2025, Wang2026}. In prior work in which ablations were performed, explicit semantic information incorporation reduced error by 50\% (LookOut), 41\% (TrajForecast) and 5\% (EgoNav) compared to a representation without precomputed features \cite{Pan2025, Qiu2022, Wang2026}.

Prior studies found that participants tend to fixate gaze upon task-relevant objects in close-quarters and display observable search behavior \cite{Zheng2022, Qiu2025}. Additionally, strong correlation has been found between regions of gaze fixation and future selected footholds in hiking scenarios \cite{Matthis2018}. In GIMO (Gaze-Informed human MOtion) authors demonstrated a 38\% reduction in translation error due to inclusion of gaze for motion prediction inside a room, while EgoCogNav observed a more modest 4\% improvement \cite{Zheng2022, Qiu2025}.

\subsection{Human Motion Datasets}

A significant amount of prior work has focused on human motion prediction in crowds or large public spaces, which do not encompass the entirety of human navigation \cite{Yehia2026, Amirian2020}. These datasets are highly applicable for robots which navigate outdoors, but indoor motion prediction represents a different and understudied problem. Prior investigation of weekly strides for a cohort of adults found that 83\% of strides taken outdoors were considered steady-state with small speed or heading changes, compared to 51\% indoors \cite{Baroudi2024_2}. Furthermore, a 39\% greater variance in stride speed was observed in indoor walking, representing an inherently more challenging prediction problem \cite{Baroudi2024_1}.

There are multiple publicly available datasets which contain human motion data, but none explicitly focus on navigation between daily tasks. The Aria Everyday Activities dataset includes a wide range of everyday motion, but a significant portion of the data focuses on activities where the user is not walking or standing, and are of less interest when aiming to predict walking navigation \cite{Lv2024}. Another dataset, EgoTraj, contains Quest Pro data from 75 subjects navigating in a variety of outdoor spaces \cite{Yehia2026}. However, this data is exclusively outdoors, and primarily in situations with crowds. There are many non-egocentric datasets encompassing the motion of large numbers of pedestrians in fairly unconstrained spaces \cite{Amirian2020}. In such cases, prior work has found that social forces from pedestrian interaction are highly relevant for refining predictions \cite{Helbing1995SocialDynamics, Xue2018}. EgoNav, which has not been publicly released, included a mixture of indoor and outdoor spaces, but the data were collected from a single non-na\"{i}ve participant, did not include eye gaze, and was not strictly purpose-driven \cite{Wang2026}. 

To investigate indoor navigation, we collected a standardized dataset across a cohort of subjects who are na\"{i}ve to specific experimental goals. We intended to emulate real motion between daily tasks by providing task goals to participants in different indoor spaces in which they are familiar.

\subsection{Motion Prediction Models}

Trajectory diffusion has been identified as a powerful, sample-efficient method of generating planned trajectories from multi-modal demonstration data in robotics \cite{Chi2023}. This has been applied in robotic manipulation, autonomous vehicle motion planning, and human motion prediction \cite{Chi2023, Wang2026, Liao2024DiffusionDrive:Driving}. By training a model which produces trajectories which imitate the training set, it can also serve as a predictor of how a person will behave given some conditioning data \cite{Tevet2022HumanModel}. Diffusion can also model multi-modal distributions, allowing a model to capture divergent probable outcomes given conditioning data.

The noise estimator used in diffusion can take many forms, and token-based transformers have provided a strong method for fusing multi-modal data \cite{Zhang2022MotionDiffuse:Model, Jiang2025TransDiffuser:Driving}. In this paper, we use a simple and generalizable approach extended from these methods. Our aim is primarily to compare relative information value rather than model structure, but we also present a robust and effective motion prediction model.

\section{DATA PREPARATION}

\begin{figure*}[t]
    \centering
    \includegraphics[width=\textwidth]{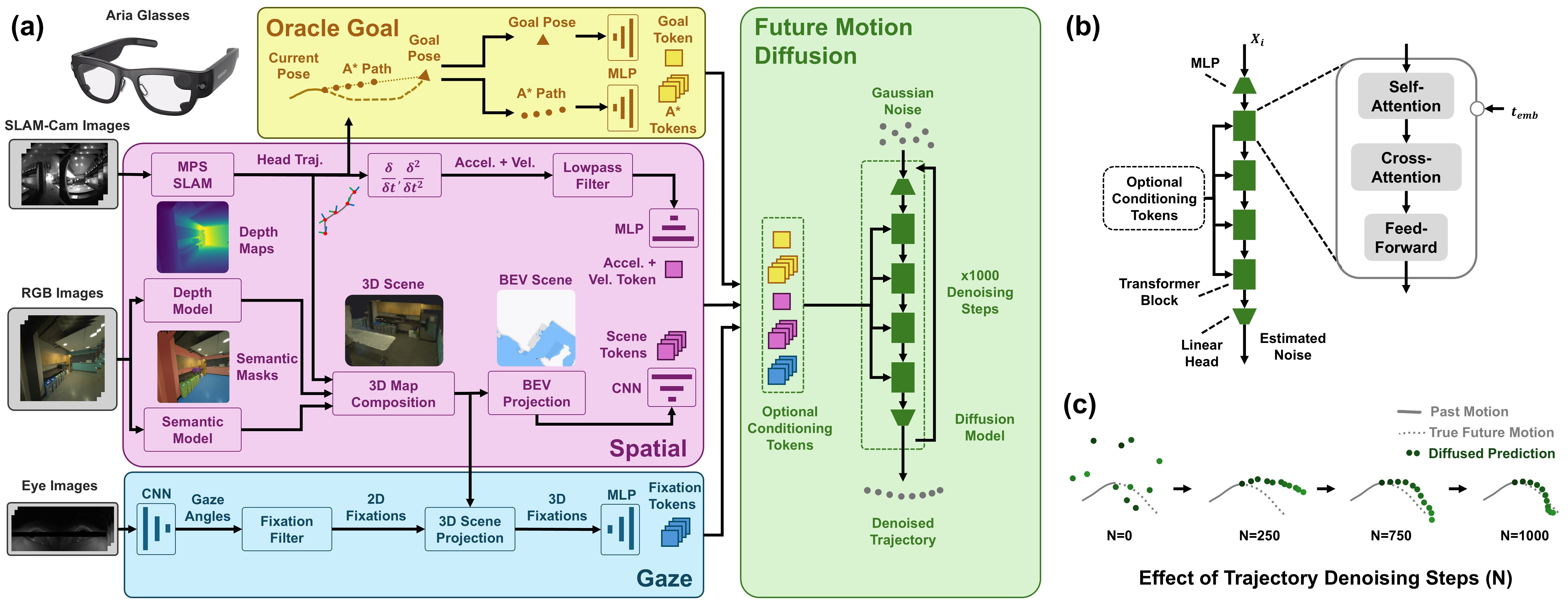}
    \caption{An overview of the data processing and motion prediction approach. (a) Describes the flow of data through the model for each of the multi-modal token sources. (b) The structure of the diffusion noise estimator, which consists of four sequential blocks with self- and cross-attention components. (c) A demonstration of the trajectory diffusion process. Further denoising steps result in smoother, more accurate trajectories.}
    \label{fig:model_overview}
\end{figure*}

This study aims to provide an assessment of relative information value in the path prediction problem, so accuracy and reliability were emphasized over a real-time implementation. A demonstration of the task-focused protocol is depicted in Fig. \ref{fig:top_down_map}, and an overview of the complete data processing pipeline is shown in Fig. \ref{fig:model_overview}a.

\subsection{Data Collection} 

\begin{figure}[t]
    \centering
    \includegraphics[width=\columnwidth]{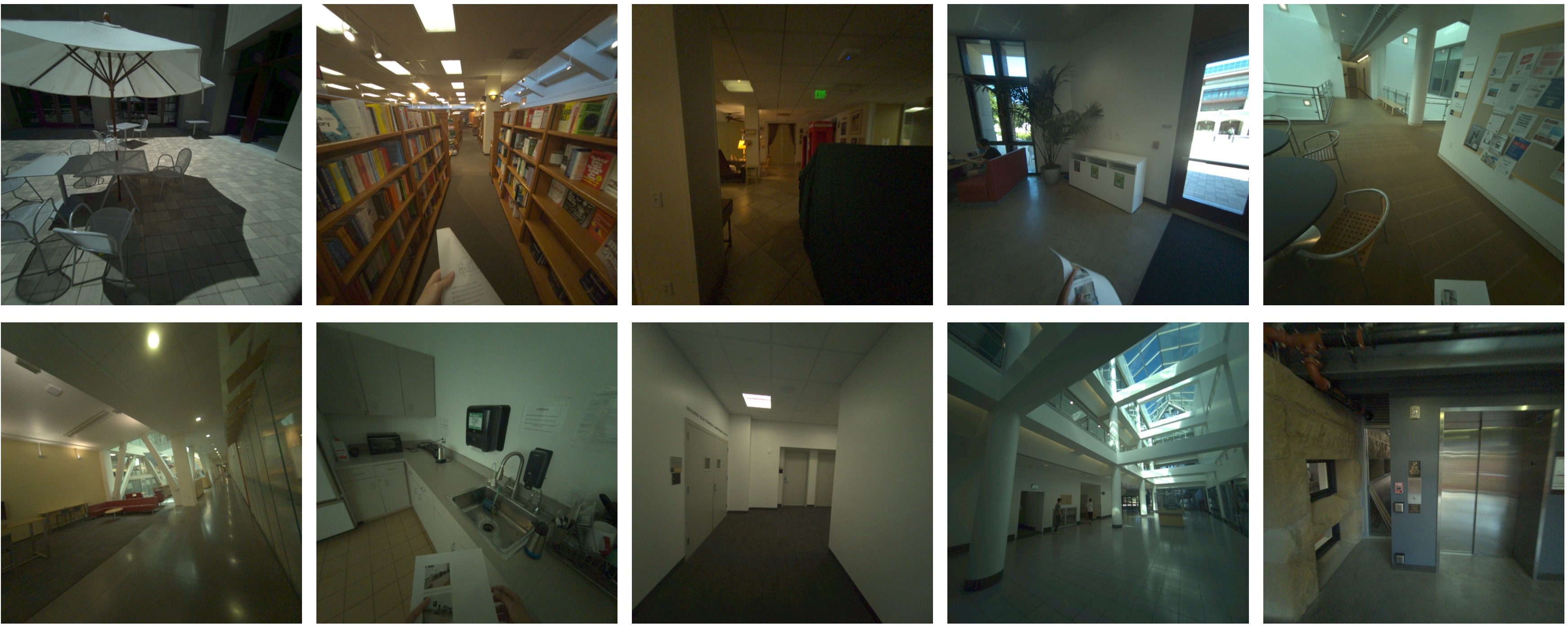}
    \caption{Example RGB frames from task-focused navigation protocol. Ten campus buildings were navigated by nine study participants.}
    \label{fig:rgb_frames}
\end{figure}

\begin{table}[h]
\caption{Summary of Collected Data}
\label{data_summary}
\begin{center}
\begin{tabular}{|l||c|c|c|c|c|}
\hline
ID & Train & Test & Total (min) & Routes & Locations \\
\hline
1 & 15.2 & 7.4 & 22.5 & 53 & A, B, D \\
\hline
2 & 18.5 & 5.1 & 23.6 & 45 & A, J \\
\hline
3 & 21.8 & 7.2 & 28.9 & 67 & A, F, H \\
\hline
4 & 18.7 & 7.0 & 25.7 & 58 & A, B, C \\
\hline
5 & 24.3 & 6.9 & 31.2 & 64 & A, D, E \\
\hline
6 & 15.4 & 7.3 & 22.7 & 59 & A, H, I \\
\hline
7 & 17.4 & 5.9 & 23.3 & 66 & A, D, H \\
\hline
8 & 21.4 & 7.8 & 29.2 & 65 & A, F, G \\
\hline
9 & 22.7 & 7.7 & 30.4 & 63 & A, B, G \\
\hline
\textbf{Total} & \textbf{175.4} & \textbf{62.2} & \textbf{237.6} & \textbf{540} & \textbf{10} \\
\hline
\end{tabular}
\end{center}
\end{table}

Data were collected using the Meta Aria Gen1 Research Kit (ARK). The ARK provides a high-resolution front-facing RGB camera, two inertial measurement units (IMUs), outward-facing mono cameras for SLAM, and inward IR cameras for gaze tracking. RGB and gaze data were collected at 20 fps, SLAM cameras at 10 fps, and IMU data at 800 hz and 1000 hz. Nine subjects without knowledge of the study goals were recruited for participation in the study, approved under a university institutional review board. Participants were asked to perform a series of simple tasks inside familiar spaces on campus. Tasks were prompted with a photo, and phrases like ``turn on the sink" or ``open the fridge", with examples shown in Fig. \ref{fig:top_down_map}. Demonstrative image frames from the subjects as they completed the tasks in different campus buildings are shown in Fig. \ref{fig:rgb_frames}. Subjects were asked to complete sets of these tasks in different orders, to provide a variety of different paths inside the same space. A train-test split of approximately 3:1 was applied among collected trials to create an isolated validation set of unseen trials for each study participant, summarized in Table \ref{data_summary}. Different task set orders were used for each participant, so that each route between two tasks was unique.

All subjects completed a single identical set of four tasks in location A, to serve as a standardized baseline for evaluation. These trials were specifically excluded from the training set for all presented models.

In total this approach produced 237.6 minutes of navigation data from a set of subjects with a wide spread of body types and preferred speeds. This spanned ten university buildings, and a total of 540 completed tasks. 

The subject cohort was comprised of nine female individuals, who were 24.6$\pm$3.6 years of age, an average height of 165$\pm$8 cm, and an average weight of 69$\pm$23 kg.

\subsection{Semantic Segmentation}

To extract further meaning from the camera data with relevance to the path prediction problem, the SAM3 segmentation model was used \cite{Carion2025}. SAM3 was used to produce a pixel-by-pixel label mask for each image frame with prompts: ``Wall.", ``Floor.", ``Person.", ``Door.", ``Stairs.", ``Appliance.", ``Seat.", ``Table.", ``Board.", ``Trash can.", ``Sink.". These labels were then incorporated into the 3D voxel representation detailed in the following section, as shown in Fig. \ref{fig:3d_example}.

\subsection{Head Pose and Dynamics} 

Head pose over time was estimated using Meta-provided Aria Machine Perception Services (MPS) \cite{Engel2023}. This yielded high accuracy metric camera pose estimates, but the exact method of estimation is not publicly disclosed. From head position and orientation data, velocity and acceleration were computed for the camera frame to be used in motion prediction. A simple derivative was taken for position and Euler angle orientation data, and causal lowpass filters of 5 hz were applied for linear velocity, angular velocity, linear acceleration, and angular acceleration to remove high frequency noise in each stream. This frequency was found to be a balance between introduced time delay, and removal of high frequency noise.

\subsection{3D-Reconstruction}

\begin{figure}[t]
    \centering
    \includegraphics[width=\columnwidth]{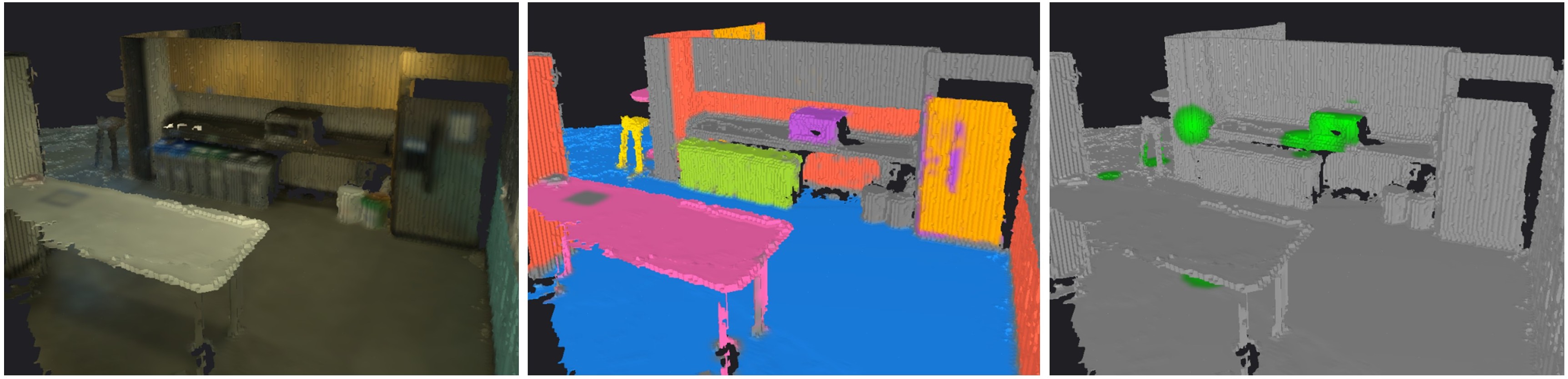}
    \caption{Layers of the 3D voxel meshes reconstructed from egocentric vision data. Color, semantic labels, and gaze fixation layers are shown. The participant was instructed to ``Open the microwave in the kitchenette".}
    \label{fig:3d_example}
\end{figure}

\begin{figure}[t]
    \centering
    \includegraphics[width=\columnwidth]{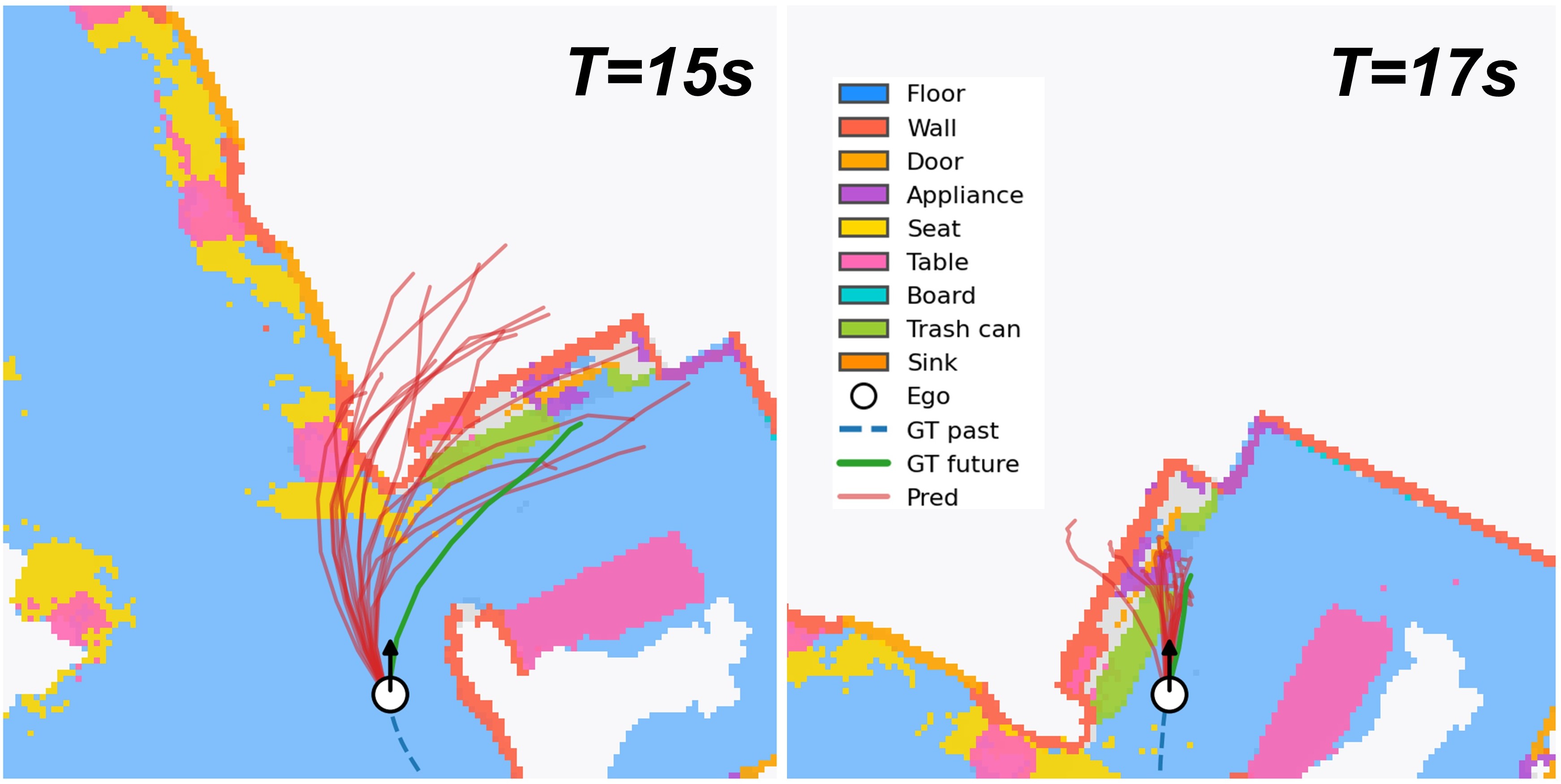}
    \caption{Example BEV frames of the kitchenette scene depicted in Fig. \ref{fig:3d_example}, with example diffusion predictions from model AVOSG.}
    \label{fig:BEV_example}
\end{figure}

To create a reconstruction of the environment for the purpose of creating simulated views, the Pi3 structure-from-motion (SfM) model was used \cite{Wang2025pi3:Learning}. Pi3 uses a batch of images with common features to create an unscaled dense 3D reconstruction and estimated camera poses. This approach was needed to acquire 3D information, because the first generation ARK does not have depth cameras. Using Pi3, reconstructions and sets of camera poses were produced for each sequential batch of 120 frames. Then, these camera poses were fit to the MPS produced trajectory data, providing a scale and common frame for the Pi3 depth images. Depth point observations were added to a 3D voxel-based representation. Sample reconstructions are shown in Fig. \ref{fig:top_down_map} and Fig. \ref{fig:3d_example}.

Each voxel in the 3D representation included an average RGB value, an observation count, an occupancy confidence value, and counts for observed semantic labels. A large representation of all observed static voxels above a certain confidence level was used instead of a point cloud to produce cleaner reconstructions and reduce memory requirements. Objects labeled with semantic label ``Person." by SAM3 were instead stored in a time-separated representation, which could then be projected at the correct timestamps. This approach was an effective way to include moving pedestrians as a special form of obstacle, but did not include the velocity of other agents.

The 3D reconstruction around the user was then projected into a simulated birds-eye-view (BEV) at each frame, an example of which is shown in Fig. \ref{fig:BEV_example}. Using only the observations made by the egocentric camera compiled in the reconstruction, a 128x128 2D representation was computed. This BEV included binary space occupancy for the area around the subject's knees (to represent walls and obstacles), and binary semantic label layers. This provided easily interpretable information about obstacles and the semantic meaning of spaces. The view used was egocentric and centered such that the subject appears at the bottom of the region, so that the future terrain is visible. This view spanned a total of 10x10 meters.

\subsection{Eye Gaze Fixation}

To estimate 3D gaze points, the Meta Eyetracking model was used to acquire gaze angle estimates. The vector formed from gaze pitch and yaw angle were then projected to the 3D voxel representation to determine the 3D scene point. This approach was used because the provided gaze tracking MPS tool was found to be inaccurate past a few meters.

Once gaze points were estimated, a gaze fixation filter was used to remove points which were considered saccades and to isolate true fixations. This provided estimates for the amount of time a fixation lasted before gaze shifted focus. Fixations shorter than 60 ms were removed, and sequential fixations with a gap less than 75 ms and pitch and yaw difference of 5 degrees were merged. This produced a set of XYZ fixations with associated durations and midpoint times.

\subsection{Oracle Goal Information}

With the aim of assessing the relevance of intent information in path prediction, task goal position was extracted from future ground truth data. For each trial, A* was used to compute the shortest valid path through the complete 3D mesh between task start and end. A* is a commonly-used algorithm to find the shortest path between two points in a graph, but does not necessarily reflect real energy-minimizing human paths \cite{Brown2021}. When the start or end of a route was infeasible due to low confidence in the 3D mesh, the closest feasible trajectory point was used for A* planning, and a straight line was drawn to close the gap. This was deemed a suitable approximation after inspection of all intent trajectories. 

To accommodate for frequent situations in which the subject did not take the shortest or optimal path, A* was re-planned every five seconds. In this format, only the most recently planned A* trajectory was used as an information input.

To translate trajectories into model inputs, the closest intent trajectory point to each egocentric position was identified. Assuming a baseline speed of 1.3 m/s along the intent trajectory, ten evenly-spaced points were sampled with a horizon of 5s from the egocentric position.

\section{APPROACH TO MOTION PREDICTION}

A diffusion model was used to generate predicted trajectories by finding the average trajectory produced by sampling a conditioned diffusion distribution. The diffusion model used was conditioned on different types of information depending on the ablation performed. Data in the test set was not used during training or tuning of model parameters; only for evaluation.

Model ablations were conducted in order of increasing system complexity. Acceleration data can be gathered from a simple IMU (A), while adding velocity data represents a further cost due to the need for odometry (AV). Adding and maintaining a 3D scene model (AVO), and incorporating semantic information (AVOS) also represent significant leaps in computational demand. Gaze data (G) is viewed as a separate and optional subsystem, and so ablations are conducted with gaze at each level of complexity. Abbreviations are A - Acceleration, V - Velocity, O - Occupancy, S - Semantic, G - Gaze. These results are compared to a constant velocity baseline (abbreviated C), which assumes that the mean past second of body velocity will continue over the prediction horizon.

\subsection{Model Structure}

The denoiser used in this study was a cross-attention based transformer, shown in Fig. \ref{fig:model_overview}b. The core of the denoiser consists of four identical blocks with different weights, each containing sequential self-attention, cross-attention, and linear head blocks. Self-attention is used to provide correspondence between sequential future trajectory points, while cross-attention is used to model relationships between conditioning data and future motion data \cite{Liao2024DiffusionDrive:Driving, Zhang2022MotionDiffuse:Model}. Length 128 tokens were computed from the input data and used to predict future motion. This model structure was selected to allow for significant generality in the included information. Each information source was used to compute tokens containing relevant information, which are used in the cross-attention blocks.

A 3-layer CNN was used to produce 64 (8x8) scene tokens from the BEV representation, such that each token had an approximate footprint of ~1.5 meters. A single token was computed from the acceleration and/or velocity at the timestep used for evaluation. Gaze fixations were processed with a 3-layer MLP which converted gaze fixation x, y, z, duration, and time since fixation to a single token per fixation. Thus there were a variable number of gaze fixation tokens used at any given time. For route or goal information, individual points were converted to tokens through 2-layer MLPs. Specific model parameters will be provided in the repository released with acceptance.

For evaluation, 16 predicted trajectories for a 5s time horizon were generated from a full 1000 denoising steps, as shown in Fig. \ref{fig:model_overview}c. Best-of-N (BoN) metrics are often used in motion prediction literature, but due to inability to select the best trajectory in practice, it is an impractical measure of real system performance. Thus, in this work we compute the mean generated trajectory, sacrificing multi-modality but providing a more useful error metric. All ablated models were trained for 50 epochs, and evaluations were performed on the final checkpoint. Evaluation loss was consistent by 50 epochs for initial trained models.

\subsection{Error Computation}

To determine the relative value of different information sources across ablations, we use average displacement error (ADE) for the predicted trajectory distribution, which is equivalent to the point-wise mean absolute error between predicted and actual trajectories. Error distributions exhibit substantial skew across test cases due to a long tail toward higher error, so we report median ADE to provide a representative example of error.

Reporting accuracy in human motion prediction is highly dataset-dependent. A set of trajectories which primarily consist of straight-line walking is far easier to estimate than more complex activities. For this reason, we separate results into four summary metrics, in which the subject moved at steady-state, turned, accelerated, or decelerated. In alignment with prior work, steady-state is defined as non-turning and non-accelerating/decelerating, using cutoffs of a speed change of 0.3 m/s, and of a heading change of 20 degrees \cite{Baroudi2024_1, Baroudi2024_2}. This provided further insight into how different sources of information improve the predictive ability of the model, and avoided the dominance of the easiest case of consistent velocity. 

For the results reported in this paper, every 20th frame in the evaluation set was used as a test case. This approach yielded 1321 steady, 1547 curved, 437 acceleration and 790 deceleration test cases.

\section{ABLATION RESULTS}

With the inclusion of inertial, spatial, semantic, and gaze information, we observed a 42\% improvement from the constant velocity baseline. In the following section, we discuss how each piece of information contributed to this reduction.

\begin{figure*}[t]
    \centering
    \includegraphics[width=\textwidth]{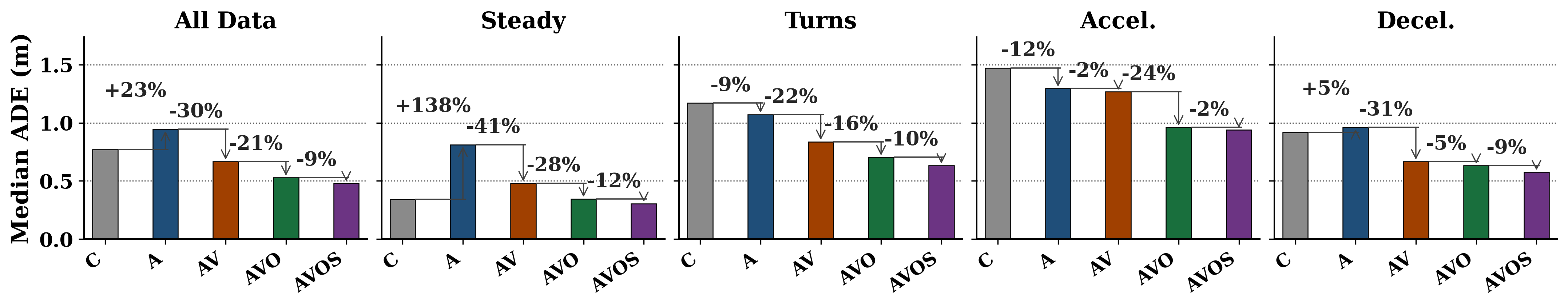}
    \caption{Relative performance of models with different information levels, across categories of motion, without gaze. Velocity and occupancy information provided the largest relative error reduction, while semantic information provided a smaller benefit. Abbreviations are C - Constant Velocity, A - Acceleration, V - Velocity, O - Occupancy, S - Semantic.}
    \label{fig:bar_chart_no_gaze}

    \centering
    \includegraphics[width=\textwidth]{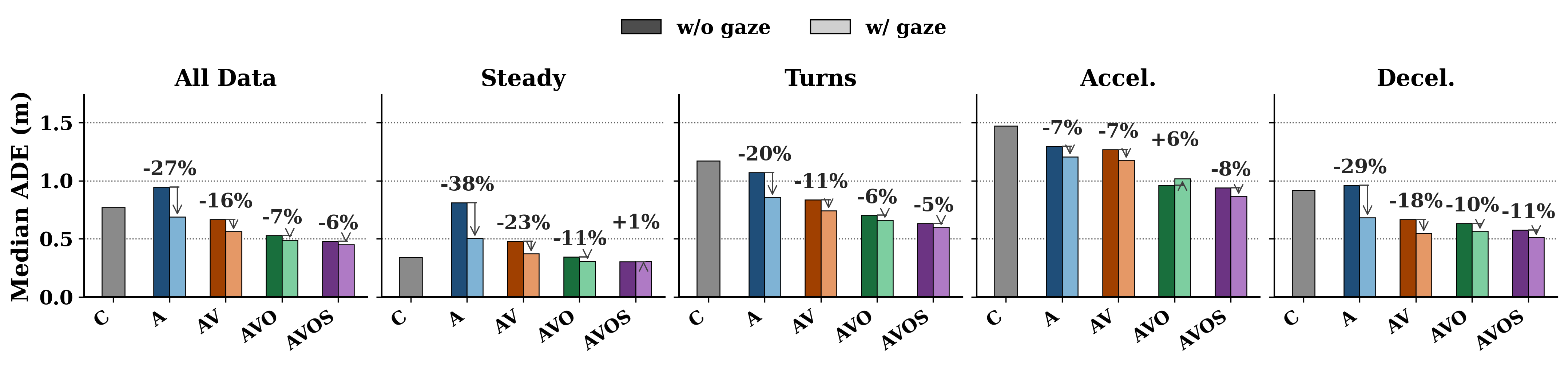}
    \caption{The effect of introducing gaze to an equivalent model without gaze for each level of model complexity. Gaze consistently improves prediction accuracy in all models and motion categories, with the exception of AVOS vs AVOSG acceleration.}
    \label{fig:bar_chart_all}
\end{figure*}

\subsection{Inertial and Scene Relevance}

As shown in Fig. \ref{fig:bar_chart_no_gaze}, providing current velocity information (AV) produced a 30\% reduction in overall error compared to providing only acceleration data (A). In non-steady cases, the predictive model (AV) outperformed the constant velocity baseline (C), but displayed poorer accuracy in  steady-state cases in which velocity and heading did not change significantly. Results shown in Fig. \ref{fig:bar_chart_no_gaze} and Fig. \ref{fig:bar_chart_all} suggest that the discrepancy is related to the model's confidence in space in front of the ego position; steady error significantly improved when either gaze (AVG) or occupancy (AVO) is included. An overly cautious model tends to underestimate motion at preferred walking velocity in case a stop or turn occurs, and in turn outperform constant velocity in situations where motion changes occur. Model AV lacks any form of obstacle information, and represents just current motion, resulting in conservative predictions.

Occupancy information produced error reductions in all categories, to a similar relative magnitude as the inclusion of velocity data. AVO provided sufficient information to match the performance of the constant velocity baseline for steady walking, while also providing large improvements of 30-40\% in more challenging cases. Semantic information (AVOS) provided a further reduction of error in all categories except acceleration, but the improvement percentage of 9\% was notably lower than inclusion of either velocity or occupancy. Prior work has demonstrated greater utility of semantic information, but this benefit was primarily in crowd scenarios which were less common in this dataset \cite{Pan2025}. These results suggest that while inclusion of inertial and occupancy information are critical, semantic information provides greater benefits in specific conditions.

\subsection{Gaze Fixation Relevance}

\begin{figure}[t]
    \centering
    \includegraphics[width=\columnwidth]{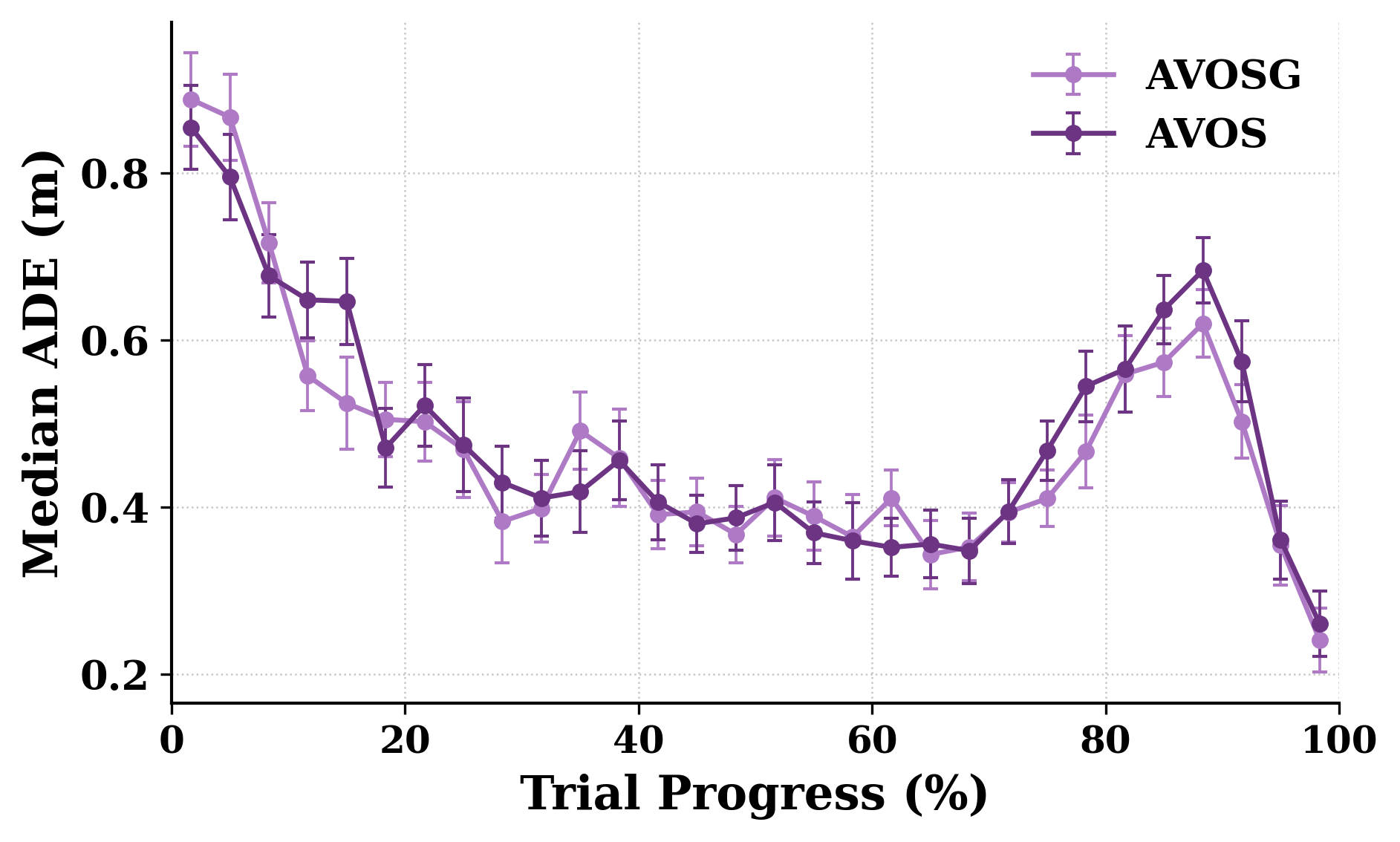}
    \caption{A comparison of median ADE by trial progress for AVOS and AVOSG. Error reduction due to gaze inclusion is largest closer to the end of the trajectory. The x-axis is normalized by the percentage temporal progress through the trial. Error bars are standard deviation of ADE.}
    \label{fig:gaze_progress}
\end{figure}

Gaze fixation provided general utility for all models, and was found to particularly improve deceleration prediction. In Fig. \ref{fig:bar_chart_all}, model AG is shown to reduce deceleration prediction error by 29\% compared to  model A. This was an effect comparable to inclusion of current velocity, which was crucial across all prediction categories. This also indicated that in many task-focused navigation scenarios, gaze alone provided sufficient information to anticipate a stop. The protocol is task-focused, so participants tended to focus on the task as they approached. This effectively provided a metric error vector to the task goal once the end is in view, evident in Fig. \ref{fig:gaze_progress}. Notably, gaze information also reduced error by 5-10\% in all non-steady categories when added to AVOS, demonstrating that gaze fixation offered further utility than just a stop indicator.

As shown in Fig. \ref{fig:bar_chart_all}, gaze fixation produced a smaller improvement in error magnitude when occupancy was already included (16\% vs 7\%). Overall error in AVO was also only 6\% lower than in AVG. These results suggest that gaze provided some metric spatial information to the model which was redundant with the spatial information which occupancy provided. Because gaze provided a collection of metric points, it could be viewed as an extremely sparse occupancy map. In this work, 2D gaze fixation vectors are mapped to 3D using collisions with the constructed scene, but prior work has demonstrated that depth alone can be used to extract sparse depth maps \cite{Kuo2018DepthGaze}. Thus, an accurate gaze tracking system which includes localization for velocity data poses a lightweight but effective information set, without the computational requirements associated with scene construction or semantic labeling. 

\subsection{Known Intent Relevance}

\begin{figure}[t]

    \centering
    \includegraphics[width=\columnwidth]{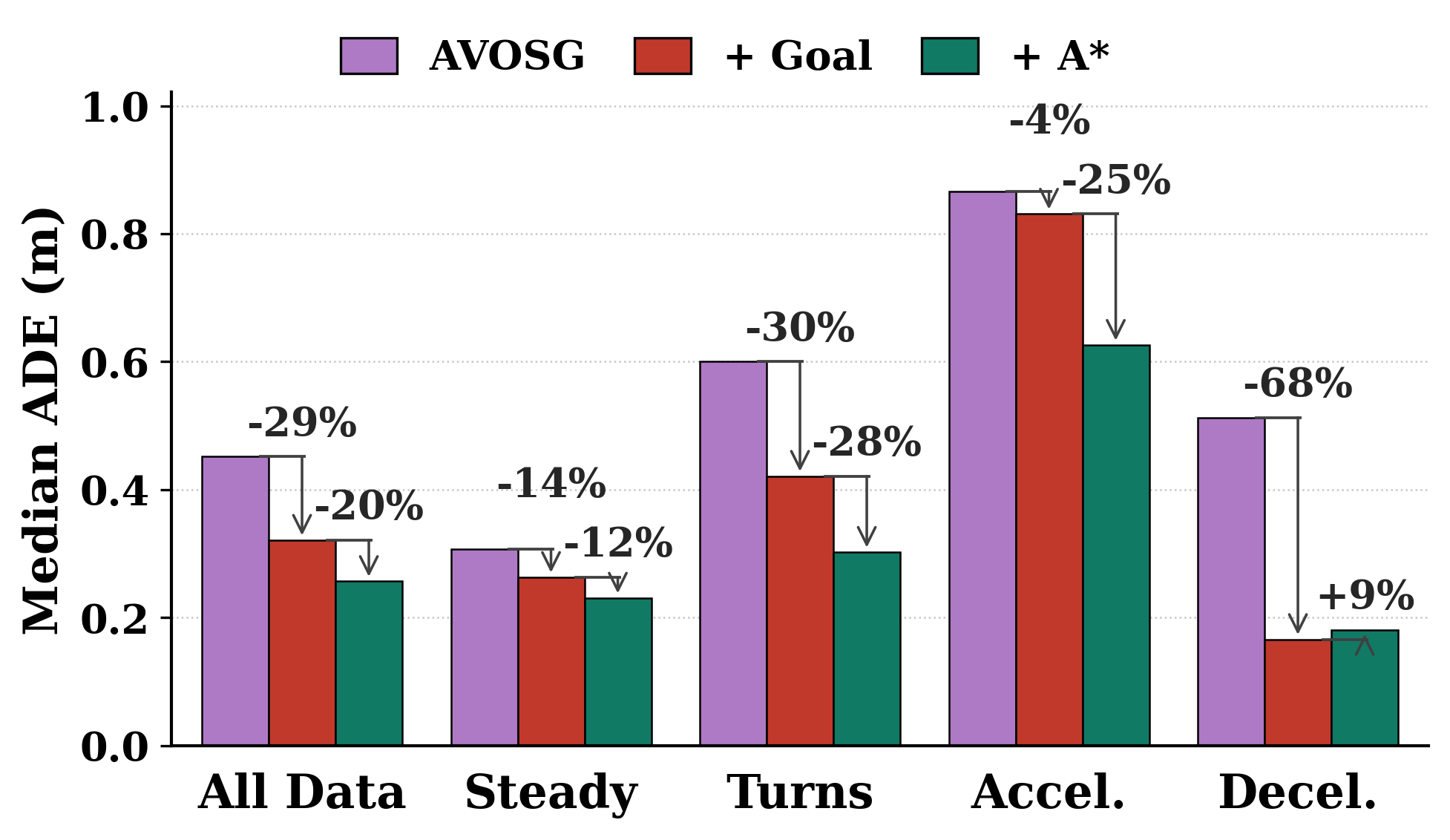}
    \caption{The impact of including explicit goal information in the best-performing ablated model (AVOSG). Goal information provides large benefits in turning and deceleration, but has limited impact on acceleration prediction accuracy.}
    \label{fig:bar_chart_known_goal}

    \centering
    \includegraphics[width=\columnwidth]{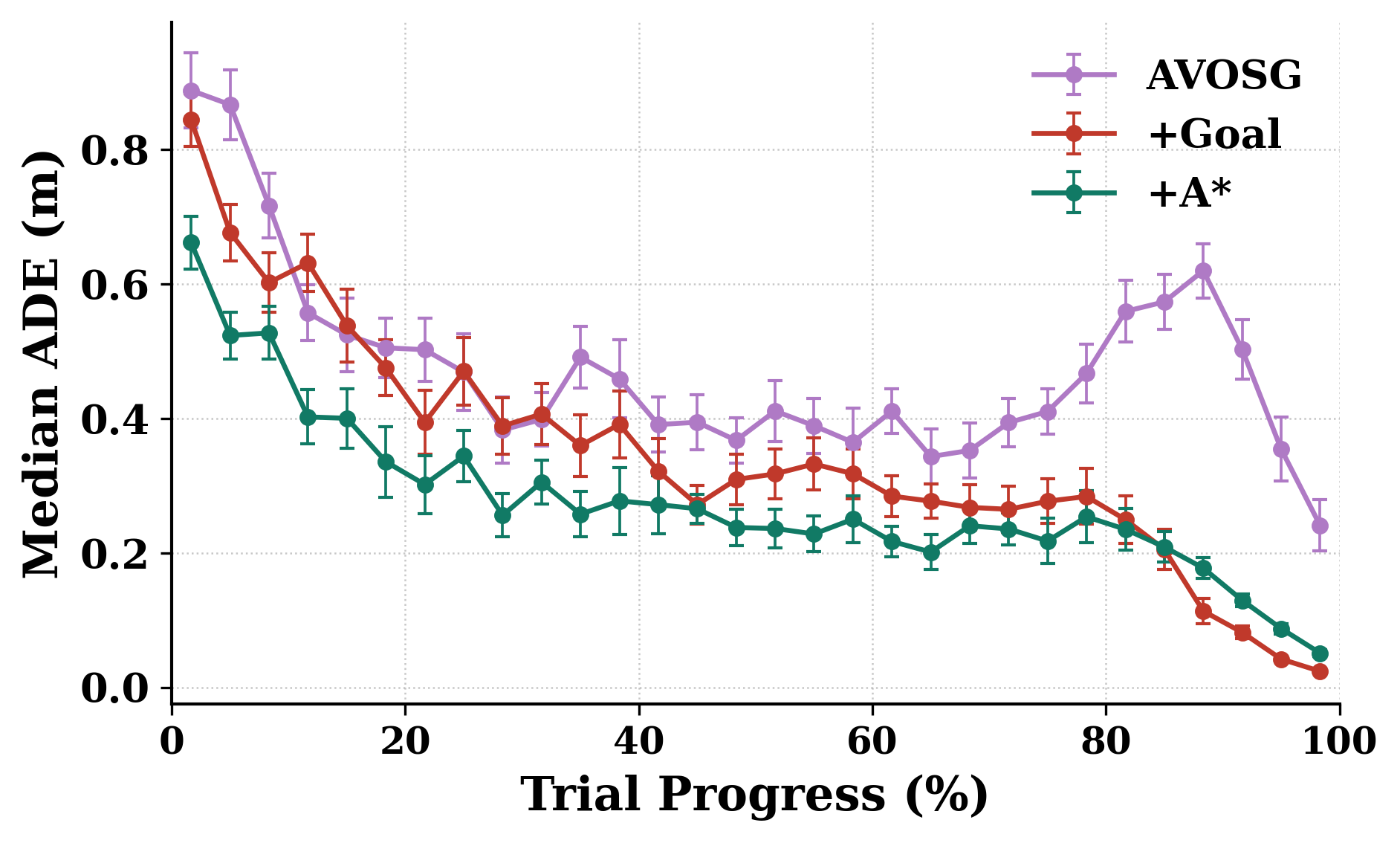}
    \caption{A comparison of median ADE by trial progress for AVOSG, AVOSG + goal point, and AVOSG + A*. The x-axis is normalized by the percentage temporal progress through the trial. Error bars are standard deviation of ADE.}
    \label{fig:goal_progress}
\end{figure}

As shown in Fig. \ref{fig:bar_chart_known_goal} and Table \ref{ablation_known_goal}, providing an error vector to the task goal position reduced prediction error by 68\% in deceleration test cases. Deceleration occurred at the end of each trial when the goal was nearby, meaning that the goal position provided an explicit stopping point which was easy for the model to interpret, visible in Fig. \ref{fig:goal_progress}. Providing goal information also improved performance in turning cases by 30\%, suggesting that ambiguity was reduced by providing explicit goal direction information. Error reductions in acceleration and steady walking were more modest, but these test cases also benefited somewhat from reduced ambiguity.

When a section of the A* shortest path to the trial end was provided, error was further reduced in all categories except deceleration. As shown in Fig. \ref{fig:bar_chart_known_goal}, this was most prominent in turns (28\%) and acceleration (25\%), which are both cases where instantaneous direction information is crucial. Cases in which subjects turned unexpectedly and moved towards the next task from rest were common in the collected data, and the shortest path to goal reduces direction ambiguity in these cases. Timing of motion onset was still ambiguous, thus acceleration prediction remains challenging. These observed improvements indicate that ambiguity remains in the user's intended route.

\begin{table}[h]
\caption{Ablation Results, Known Goal. Median ADE reported (m)}
\label{ablation_known_goal}
\begin{center}
\begin{tabular}{|l||c|c|c|c|c|}
\hline
Model & All $\downarrow$ & Steady $\downarrow$ & Turn $\downarrow$ & Accel. $\downarrow$ & Decel. $\downarrow$ \\
\hline
C & 0.77 & 0.34 & 1.17 & 1.47 & 0.92 \\
\hline
AVOSG & 0.45 & 0.31 & 0.60 & 0.87 & 0.51 \\
\hline
+ Goal & 0.32 & 0.26 & 0.42 & 0.83 & \textbf{0.17} \\
\hline
\textbf{+ A*} & \textbf{0.26} & \textbf{0.23} & \textbf{0.30} & \textbf{0.63} & 0.18 \\
\hline
\end{tabular}
\end{center}
\end{table}

Fig. \ref{fig:goal_progress} demonstrates that for AVOSG without future information, both the start and end of each trial was challenging to predict due to motion initiation and stopping respectively. Goal information significantly reduced error at the end of trials, and provided a small reduction at the beginning. Fig. \ref{fig:bar_chart_known_goal} shows that acceleration events are more challenging to predict than any other category, even when goal information is provided. This suggests that another approach is warranted to anticipate motion initiation.

\subsection{Variation in Human Motion}

Error remained even when route was known, due to a combination of natural variance in paths taken and unmodeled factors which affect path structure. An estimate of path variance was calculated from location A data, in which each subject completed an identical set of four routes. Task completion constrained the start and end of each subject's four routes, but not the exact path and speed which they used to complete the tasks. To resolve coordinate frame differences, Kabsch-Umeyama was used to find the closest possible alignment of the trajectories from each trial, and trajectories were then resampled to 400 points to remove differences in progress rate, similar to computing the Fr\'{e}chet distance. 

For the eight subjects who took an identical route, the average RMSE from the mean trajectory was 0.17 m, which increased during turns (0.19 m) and acceleration (0.20 m). These results suggest that minimal possible prediction error was not zero, due to natural variation in human motion, and that departures from steady state resulted in even higher variability. This also demonstrated that there were unmodeled factors in the AVOSG + A* model, indicating that further improvements are possible. However, further investigation is warranted to extend this analysis.

\subsection{Generalization to Unseen Subjects}

To evaluate model generalization ability, nine AVOSG models were trained, each excluding one subject. Each model was evaluated only on the subject which did not appear in the training set, and results were included in the same median computation. Thus, the evaluation set was identical between all-subject AVOSG and subject-exclusionary AVOSG. Furthermore, to remove the risk of information leakage from familiar locations included in the training set, evaluation was only performed on the standardized introduction task set from location A. 

\begin{table}[h]
\caption{Ablation Results, Subject Holdout. Median ADE reported (m)}
\label{ablation_subject_holdout}
\begin{center}
\begin{tabular}{|l||c|c|c|c|c|}
\hline
Model & All $\downarrow$ & Steady $\downarrow$ & Turn $\downarrow$ & Accel. $\downarrow$ & Decel. $\downarrow$ \\
\hline
\textbf{No Holdout}& \textbf{0.61} & \textbf{0.29} & \textbf{0.84} & \textbf{0.95} & \textbf{0.67} \\
\hline
w/ Holdout & 0.64 & 0.32 & 0.85 & 0.98 & 0.76 \\
\hline
\end{tabular}
\end{center}
\end{table}

Subject exclusion increased median ADE across all categories by only 4.9\%, with specific results visible in Table \ref{ablation_subject_holdout}. Based on these results, low subject-specific dependence was observed, suggesting that our model generalized well to new subjects but did benefit from prior subject data.

\section{CONCLUSION}

This work presents an assessment of relative information value in human motion prediction for use in human-robot interaction. In alignment with prior work, inertial and occupancy information significantly reduced error in motion prediction.  In task-focused motion, we find that gaze is a valuable signal to both indicate intent to decelerate before it occurs, and to provide basic spatial information. Notably, these results are based on natural gaze behavior from subjects who were unaware of experimental goals. Using gaze as an active control signal to disambiguate uncertain situations is a promising direction of future work.

Even with gaze data, ambiguous situations are common in human motion, and without higher-level knowledge of task goals or individual routines, accurate long-horizon motion prediction is a major challenge. Analysis of models with oracle task and route information demonstrated that intent inference is still a challenge, even with significant other sources of information. Particularly, the challenge of predicting the timing of motion initiation is highly relevant for indoor scenarios and seemingly infeasible with vision data alone. This warrants further work, either through the use of low-latency controllers which quickly detect motion or by incorporating new sensing modalities, like electromyography.

Semantic labels were generally informative, but did not produce the same magnitude of improvement as prior outdoor crowd-focused work (LookOut). This suggests that for indoor, task-focused navigation, supplying semantic information is less informative than in outdoor crowd scenarios. Future work could incorporate the use of semantic label filtering by isolating task-relevant labels, instead of including all possible task objects.

In this work we do not directly present a real-time prediction system. However, prior work has already demonstrated that using online SLAM and introducing a DDIM/DDPM structure for diffusion makes real-time diffusion predictions feasible \cite{Wang2026}. We next aim to use our predictive model with a walking exoskeleton to anticipate user speed changes online. This will provide better estimates of ankle torque timing in transitionary steps, increasing user stability while providing walking assistance.

\addtolength{\textheight}{-12cm}   





\bibliographystyle{IEEEtran}
\bibliography{citations}

\end{document}